%% file: emnlp2018.tex
\documentclass[11pt,a4paper]{article}
\usepackage[nohyperref]{emnlp2018}
\usepackage{times}
\usepackage{latexsym}
\usepackage{xcolor}
\usepackage{booktabs}
\usepackage{url}

\usepackage{bbm}
\usepackage{multirow}
\usepackage{booktabs}
\usepackage{url}
\usepackage{xcolor}
\usepackage{graphicx}
\usepackage{tabularx}
\usepackage{verbatim}
\usepackage{enumitem,kantlipsum}
\usepackage{multirow}
\usepackage{amsmath}
\usepackage{amssymb}

\newcommand{\boldcy}[1]{{\textit{\color{purple}{#1}}}}
\newcommand{\boldred}[1]{{\textbf{\color{red}{#1}}}}

\aclfinalcopy 

\usepackage{pifont}

\title{Neural Metaphor Detection in Context}

\author{Ge Gao,$^{\diamondsuit}$ Eunsol Choi,$^{\diamondsuit}$ Yejin Choi,$^{\diamondsuit\dagger}$ and Luke Zettlemoyer$^\diamondsuit$\\
\hspace{4mm} University of Washington$^{\diamondsuit}$\\  
Allen Institute for Artificial Intelligence$^{\dagger}$
\vspace{2mm} \\
\texttt{\{ggao,eunsol,yejin,lsz\}@cs.washington.edu} \\}

\date{}

\begin{document}
\maketitle

\begin{abstract}
We present end-to-end neural models for detecting metaphorical word use in context.
We show that relatively standard BiLSTM models which operate on complete sentences work well in this setting, in comparison to previous work that used more restricted forms of linguistic context. 
These models establish a new state-of-the-art on existing verb metaphor detection benchmarks, and show strong performance on jointly predicting the metaphoricity of all words in a running text. 
\end{abstract}

\input{intro}
\input{task.tex}

\input{model}
\input{data}

\input{experiment}
\input{related}
\input{conclusion}
 \section*{Acknowledgments}
We thank the anonymous reviewers for their insightful comments. 
This work was supported in part by
the NSF (IIS-1714566 and IIS-1252835), the ARO (W911NF-16-1-0121), the DARPA CwC program through ARO (W911NF-15-1-0543), and gifts from Google and Facebook.
 
\bibliography{emnlp2018}
\bibliographystyle{acl_natbib_nourl}

\end{document}

%% file: intro.tex
\section{Introduction}
Metaphors are pervasive in natural language, and detecting them requires challenging contextual reasoning about whether specific situations can actually happen.~\cite{Lakoff}. 
For example, in Table~\ref{fig:examples}, ``examining'' is metaphorical because it is impossible to literally use a ``microscope'' to examine an entire country. 
In this paper, we present end-to-end neural models for metaphor detection, which can learn rich contextual word representations that are crucial for accurate interpretation of figurative language.
 
In contrast, most previous approaches focused on limited forms of linguistic context, for example by only providing SVO triples such as (car, \textbf{drink}, gasoline) to the model~\cite{Shutova:16BlackHA,Tsvetkov2013CrossLingualMD,Rei2017GraspingTF,Bulat2017ModellingMW}. While the verbal arguments provide strong cues, providing the full sentential context supports more accurate prediction, as seen in Table~\ref{fig:examples}. 
Even in the few cases when the full sentence is used~\cite{koper2017improving,Turney2011LiteralAM,Jang16} existing models have used unigram-based features with limited expressivity.

We investigate two common task formulations: (1) given a target verb in a sentence, classifying whether it is metaphorical or not, and (2) given a sentence, detecting all of the metaphorical words (independent of their POS tags). We find that relatively standard architectures based on bi-directional LSTMs~\cite{hochreiter1997long} augmented with contextualized word embeddings~\cite{Peters2018DeepCW} perform surprisingly well on both tasks, even with modest amount of training data. 
We improve the previous state-of-the-art by 7.5 F1 on the VU Amsterdam Metaphor Corpus (VUA) for the sequence labeling task~\cite{Steen:10}, by 2.5 F1 on the VUA verb classification dataset, and by 4.9 F1 on the MOH-X dataset~\cite{Mohammad:16}. 
Our code is publicly available at \url{https://github.com/gao-g/metaphor-in-context}.

\begin{table}

\begin{center}
\begin{tabular}{p{210pt}}
\toprule
The experts started \boldred{examining} the Soviet Union with a microscope to study perceived changes. \\ \hline
Rockford teachers are honored for saving a \boldcy{drowning} student.\\\hline
You're \boldred{drowning} in student loan debt. \\ \bottomrule
\end{tabular}
\end{center}
\caption{Metaphorical usages of the target word are bold faced, and literal usages are italicized. Full sentence context is crucial for metaphor detection.}\vspace{-10pt}
\label{fig:examples}
\end{table}

%% file: task.tex
\section{Task}
We study two task formulations. 
\vspace*{-1mm}
\begin{description}
\item[Sequence Labeling:]
Given a sentence $x_1,$$\dots,$$x_n$, predict a sequence of binary labels $l_1, \dots, l_n$ to indicate the metaphoricity of each word.
\vspace*{-1mm}
\item[Classification:]
Given a sentence $x_1,\dots,x_n$ and a target verb index $i$, predict a  binary label $l$ to indicate the metaphoricity of the target $x_i$.
\end{description}
\vspace*{-1mm}
While both formulations have been studied in previous work, it is worth noting that the sequence labeling task generalizes the classification task in that the prediction for the target verb can be extracted from the full sentence predictions. 
In addition, as will be shown in Section~\ref{sec:experiments}, we find that given accurate annotations for all words in a sentence, the sequence labeling model outperforms the classification model even when the evaluation is set up as a classification task.

%% file: model.tex
\section{Model}

Our models use a bidirectional LSTM to encode a sentence, and a feedforward neural network for classification, optimized for the log-likelihood of gold labels.

\paragraph{Sentence encoding}
For both sequence labeling and classification, we represent each token $x_i$ in the input sentence with a pre-trained word embedding $w_i$. To further encode contextual information, we also concatenate ELMo (Embeddings from Language Models) vectors $e_i$ from Peters et al.~\shortcite{Peters2018DeepCW}. 
These vectors have been shown to be useful  for word sense disambiguation, a task closely related to metaphor detection ~\cite{Birke2006ACA}.

\begin{figure}\vspace{-10pt}\centering
\includegraphics[width=0.45\textwidth]{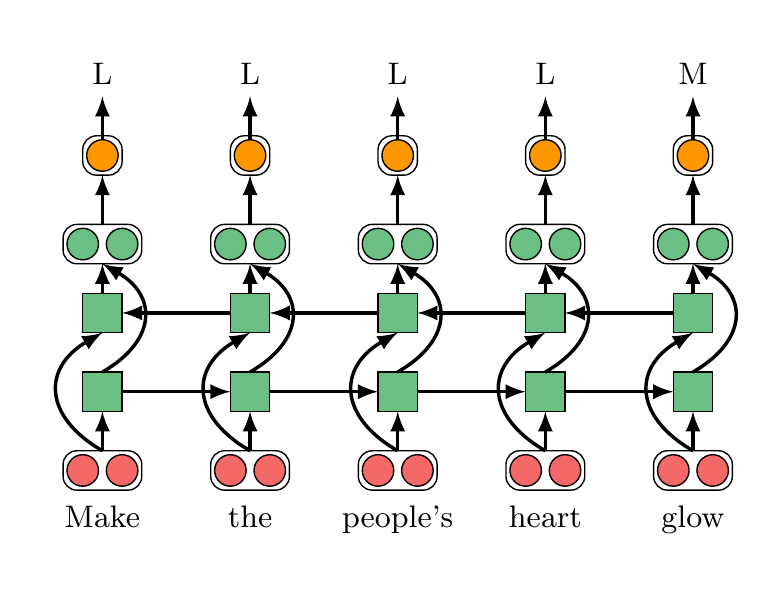}\vspace{-10pt}
\caption{A sequence labeling model for metaphor detection. Every word in a sentence is classified.}\vspace{-10pt}
\label{fig:seq}
\end{figure}

\begin{figure}\vspace{-10pt}
\centering
\includegraphics[width=0.5\textwidth]{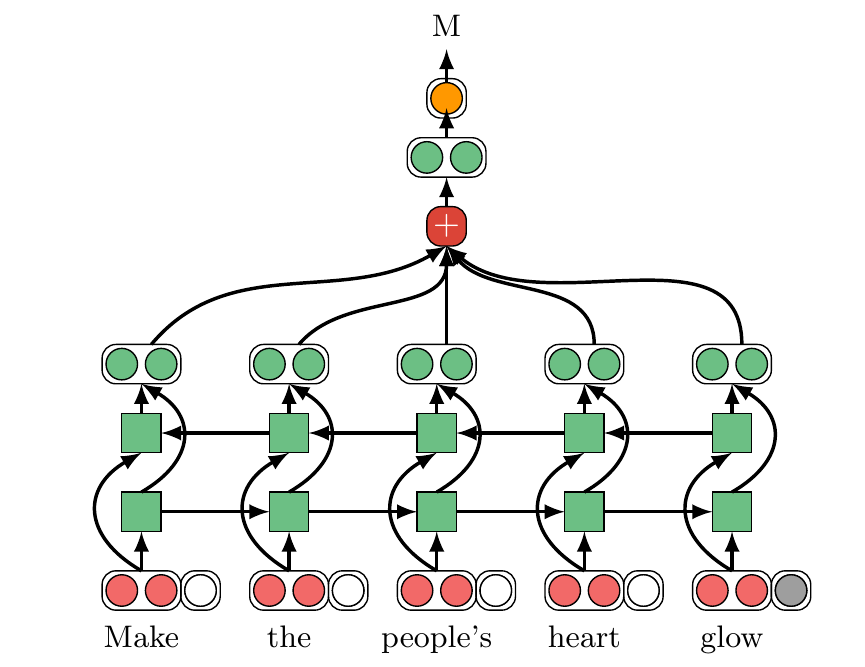}\vspace{-10pt}
\caption{A classification model for metaphor detection. Only a single word per sentence is labeled as metaphorical or literal.}\vspace{-10pt}
\label{fig:attn}
\end{figure}

\label{sec:model}

\subsection{Sequence Labeling Model}
Figure~\ref{fig:seq} shows the model architecture. 
We input the word representation $[w_i;e_i]$ to a bidirectional LSTM, 
producing a contextualized representation $h_i$ for each token. Then we use a feedforward neural network that takes $h_i$ to predict a label $l_i$ for each word $x_i$.

When the dataset does not contain annotations for every word, we make the simplifying assumption that every unannotated word is used literally. 

\subsection{Classification Model}
Figure~\ref{fig:attn} shows the model architecture. 
We concatenate an index embedding $n_i$, which indicates whether $x_i$ is the target verb. We use $[w_i;e_i;n_i]$ as an input to a bidirectional LSTM, 
producing a contextualized representation $h_i$. 

We add an attention layer by computing the attention weight $a_i$ for token $x_i$, and compute the representation $c$ as a weighted sum of LSTM output states where $W_a$ and $b_a$ are learned parameters.
\begin{align*}
    a_i &= \text{SoftMax}_i (W_a h_i + b_a)\\
    c &= \sum_{i=1}^{n} a_{i} h_{i}
\end{align*}
Finally, we feed $c$ to a feedforward network to compute the label scores for target verb.

%% file: data.tex
\section{Dataset}
We evaluate performance on a number of benchmark datasets, including two for classification (TroFi and MOH\textbackslash MOH-X) and one for tagging (VUA).\footnote{For detailed information about each dataset, please refer to original papers: TroFi~\cite{Birke2006ACA}, MOH~\cite{Mohammad:16}, VUA~\cite{Steen:10}. MOH-X refers to a subset of MOH dataset used in previous work~\cite{Shutova:16BlackHA} where verb and its argument are extracted from each sentence.}
Table~\ref{tab:data} shows statistics for the verb classification datasets. Despite being two times larger than the MOH dataset, the TroFi dataset contains only 50 unique verbs, and the larger VUA dataset contains over 2K unique verbs. The MOH dataset contains shorter and simpler sentences (example sentences in WordNet), compared to sentences in other datasets which come from resources such as news articles. The TroFi and MOH-X datasets are constructed to have higher percentages of metaphor, compared to the natural likelihood of metaphor in a running text, as seen in the VUA dataset. 

\paragraph{Classification Experiment Setup}
We perform 10 fold cross-validation on the MOH-X and TroFi datasets, following prior work. For the VUA dataset, we use the original training and test split ~\cite{Klebanov:16}, and set aside 10\% of the training set as a development set.

\paragraph{Sequence Labeling Experiment Setup}
The VUA dataset contains annotations for all words in each sentence. We divide the data into training, development, and test set following the same split for the VUA verb classification task. While the label classes are less balanced (only 11\% metaphors at the token level), this dataset is much bigger. Table~\ref{tab:seqdata} shows the data statistics.

\begin{table}
\small
\begin{center}
\begin{tabular}{l|r|r|r|r}
\toprule
& \# & \% & \# Uniq. & Avg \#  \\
& Expl. & Metaphor  &  Verb &  Sent. Len\\ \midrule
MOH-X &647 & 49\%  & 214&8.0\\
MOH &1,639 & 25\%  & 440 &7.4\\
TroFi  & 3,737& 43\% & 50  & 28.3\\
VUA &23,113 & 28\%  & 2047 &24.5 \\
\bottomrule
\end{tabular} 
\end{center}\vspace{-8pt}
\caption{Verb classification dataset statistics.   \% Metaphor refers to sentence-level percentage.}\label{tab:data}
\end{table}

\begin{table}

\begin{center}
\begin{tabular}{l|r|r|r}
\toprule
& Train & Dev & Test \\\midrule
\# Unique tokens & 13,843&7,458&7,200\\
\# Tokens & 116,622&38,628&50,175\\
\# Unique sent. & 6,323 &1,550&2,694\\
\% Metaphor & 11.2& 11.6 & 12.4\\\bottomrule
\end{tabular}
\end{center}\vspace{-8pt}
\caption{VUA sequence labeling dataset statistics.   \% Metaphor refers to token-level percentage.}\label{tab:seqdata}
\end{table}

%% file: experiment.tex
\begin{table}[t]
\small
\begin{center}
\begin{tabular}{l|r|r|r|r}
\toprule
{Model} 	& P & R & F1 & Acc.\\\midrule
Lexical Baseline& 68.6 & 45.2 & 54.5 & 90.6 \\ 
Wu~\shortcite{Wu2018NeuralMD} ensemble & 60.8 & 70.0 & 65.1 & - \\
Ours (SEQ) & \textbf{71.6} &	\textbf{73.6}	&\textbf{72.6} & \textbf{93.1}\\
\bottomrule
\end{tabular}
\end{center}\vspace{-8pt}
\caption{Performance on the VUA sequence labeling test set for all POS tags.}\label{tab:seq}
\end{table}

\begin{table}
\small
\begin{center}
\begin{tabular}{l|r|r|r|r|r}
\toprule
POS & \#  & \% metaphor & P & R & F1.\\\midrule
VERB  & \textbf{20K} & 18.1 & 68.1	 & 71.9	&69.9\\
NOUN & 20K & 13.6 &59.9	&60.8&	60.4 \\
ADP  & 13K & \textbf{28.0}&\textbf{86.8}&	\textbf{89.0}	&\textbf{87.9}\\
ADJ & 9K & 11.5 &56.1&	60.6&	58.3\\
PART & 3K & 10.1 &57.1&	59.1	&58.1\\
\bottomrule
\end{tabular}
\end{center}\vspace{-8pt}
\caption{The breakdown of performance on the VUA sequence labeling test set by POS tags. We show data statistics (count, \% metaphor) on the training set. We only show POS tags whose \% metaphor $>$ 10.}\label{tab:pos}
\end{table}

\begin{table*}
\small
\begin{center}
\begin{tabular}{l|r|r|r|r|r|r|r|r|r|r|r|r|r}
\toprule
\multirow{2}{*}{Model} & \multicolumn{4}{c|}{MOH-X (10 fold)} &  \multicolumn{4}{c|}{TroFi (10 fold)} & \multicolumn{4}{c}{VUA - Test} \\
 & P & R & F1 & Acc. &  P & R & F1 & Acc.  & P & R & F1 & Acc. & MaF1\\\midrule
Lexical Baseline & 39.1 & 26.7 & 31.3 & 43.6 &  \textbf{72.4} & 55.7 & 62.9 & 71.4 & 67.9 & 40.7 & 50.9 & 76.4  & 48.9 \\ \midrule
Klebanov (2016) & - & - & - & - &  - & - & - & - &- & - & - & -  &60.0\\
Rei (2017)&  73.6& 76.1& 74.2& 74.8 & - & - & - & - &  - & - & - & -  & -\\ 
K\"{o}per~(2017) & - & - & - &- & - & & \textbf{75.0} & - & - & - &  62.0 &-&-\\ 
Wu~(2018) \scriptsize{ensemble} &  - & - & - & - & - & - & - & - & 60.0 & \textbf{76.3} & 67.2 & -  & -\\ \midrule
CLS &{75.3}&	84.3&	\textbf{79.1}&	\textbf{78.5}&	68.7&	\textbf{74.6}&	72.0&	73.7 & 53.4 & 65.6 & 58.9 & 69.1 & 53.4\\
SEQ  & \textbf{79.1}&	73.5&	75.6&	77.2 &  70.7	&71.6	&71.1&	\textbf{74.6} & \textbf{68.2} & 71.3 & \textbf{69.7} & \textbf{81.4} & \textbf{66.4}\\
\bottomrule
\end{tabular}
\end{center}\vspace{-5pt}
\caption{Model performances for the verb classification task. Our models achieve strong performance on all datasets. The CLS model performs better than the SEQ model when only one word per sentence is annotated by human (TroFi and MOH-X). When all words in the sentence are accurately annotated (VUA), the SEQ model outperforms the CLS model.}\label{tab:main}\vspace{-8pt}
\end{table*}

\section{Experiments}
\label{sec:experiments}

\paragraph{Evaluation Metric}
We report precision, recall and F1 measure for the metaphor class as well as the overall accuracy. For the VUA dataset, we also report macro-averaged F1 score across four genres (conversation, academic writing, fiction and news).

\paragraph{Comparison Systems}
We propose a simple yet effective lexical baseline. It assigns the metaphor label if the word is annotated metaphorically more frequently than as literally in the training set, and the literal label otherwise. 
We also compare our models to previously published work, including: (1) a logistic regression classifier with features that indicate verb lemmas and the verbs' semantic class from WordNet~\cite{Klebanov:16}, (2) a neural similarity network with skip-gram word embeddings~\cite{Rei2017GraspingTF}, (3) a balanced logistic regression classifier on target verb lemma that uses a set of features based on multi-sense abstractness rating~\cite{koper2017improving}, and (4) a CNN-LSTM ensemble model with weighted-softmax classifier which incorporates pre-trained word2vec, POS tags, and word cluster features~\cite{Wu2018NeuralMD}.\footnote{The best performing model on the VUA Metaphor Detection Shared Task at the NAACL 2018 workshop on Figurative Language Processing.} 

We experiment with both sequence labeling model (SEQ) and classification model (CLS) for the verb classification task, and the sequence labeling model (SEQ) for the sequence labeling task. 

\paragraph{Implementation Details}
We used 300d GloVe vectors~\cite{Glove:14} and 1024d ELMo vectors.  We used additional 50d index embedding for the classification task. 
The LSTM module has a 300d hidden state. We applied dropout on the input to LSTM and on the input to the feedforward layer. We fine-tuned learning rate and dropout rate for each model on each dataset. We used SGD to optimize the CLS model and Adam~\cite{adam} for the SEQ model. 
We used spaCy~\cite{spacy2} for lemmatization, tokenization, and part-of-speech tagging.

\begin{table}

\begin{center}
\begin{tabular}{l|r|r|r|r}
\toprule
Model &  P & R & F1. & Acc.\\\midrule
SEQ&\textbf{68.3}&	\textbf{72.0}&	\textbf{	70.4}&	\textbf{83.5}\\
-ELMo&59.4&	64.3	&	61.7&	78.2\\
CLS&52.4&	63.0		&57.3&	74.3\\
-ELMo&52.0	&48.7&		50.8&	74.1\\
\bottomrule
\end{tabular}
\end{center}\vspace{-5pt}
\caption{Ablation study on VUA development set for the verb classification task.}\vspace{-5pt}
\label{tab:abl}
\end{table}

\paragraph{Sequence Labeling Results}
Performance on the sequence labeling task is reported in Table~\ref{tab:seq}. While prior work~\cite{Klebanov2014DifferentTS,zbal2016LearningTI} reported on the same dataset, the experiment setting is not comparable (they did cross validation on a smaller training set).\footnote{As a point of reference, their macro-averaged F1 scores were 33.25 / 50.6 respectively.} 
Our lexical baseline performs strongly in terms of precision, as some words and POS tags are almost exclusively annotated as literal. Our sequence labeling model mainly improves recall.

Table~\ref{tab:pos} reports the breakdown of performance by POS tags. Not surprisingly, tags with more data are easier to classify. Adposition is the easiest to identify as metaphorical and is also the most frequently metaphorical class (28\%). 
On the other hand, particles are challenging to identify, since they are often associated with multi-word expressions, such as ``put \textbf{down} the disturbances".

\paragraph{Verb Classification Results} Table~\ref{tab:main} shows performance on the verb classification task for three datasets (MOH-X , TroFi and VUA).\footnote{We did not compare to Shutova et al.~\shortcite{Shutova:16BlackHA} as their experiment setting is not comparable.} 

Our models achieve strong performance on all datasets, outperforming existing models on the MOH-X and VUA datasets. On the MOH-X dataset, the CLS model outperforms the SEQ model, likely due to the simpler overall sentence structure and the fact that the target verbs are the only words annotated for metaphoricity. For the VUA dataset, where we have annotations for all words in a sentence, the SEQ model significantly outperforms the CLS model. This result shows that predicting metaphor labels of context words helps to predict the target verb. We hypothesize that K\"{o}per et al.~\shortcite{koper2017improving} outperforms our models on the TroFi dataset for a similar reason: their work uses concreteness labels, which highly correlate to metaphor labels of neighboring words in the sentence. Also, their best model uses the verb lemma as a feature, which itself provides a strong clue in the dataset of 50 verbs (see lexical baseline).

Table~\ref{tab:abl} shows an ablation study on input representations (with or without ELMo vectors). Contextualized word vectors improve the performance of both models by a large margin.

\newcommand\cmark {{\ding{52}}}
\newcommand\xmark {{\ding{55}}}
\begin{table*}
\small
\begin{center}
\begin{tabularx}{16cm}{c|c|X|c}
\toprule
CLS & SEQ & Sentence & Metaphor Type  \\ \midrule
\xmark& \xmark&To \boldcy{throw} up an impenetrable Berlin Wall between you and them could be tactless. & - \\ \midrule
\xmark& \xmark &In reality you just invent a tale, as if you were \boldred{sitting} round a fire in a cave. & direct metaphor \\ \midrule
 \xmark& \xmark&So they \boldred{bought} immunity. & indirect metaphor \\ \midrule
\xmark& \xmark &During the early states of the phased evacuation the logistical problem \boldred{facing} the police was the street-by-street warning of the population to make ready for evacuation. & indirect metaphor\\ \midrule
\xmark& \cmark&There are few things worse than being \boldred{bludgeoned} into reading a book you hate. & indirect metaphor \\ \midrule
\xmark& \cmark&He thought of thick, fat, hot motorways \boldred{carving} up that land. & personification \\ \midrule
\xmark& \cmark&One might \boldcy{ask} whether motorists are ever justified in knowingly taking risks with other people's lives. & - \\ \midrule
\xmark& \cmark&The abstract talk of \boldcy{commuting} by rail or road being replaced by information technology finds a concrete expression in the idea of telecottages. &- \\ \midrule
\xmark& \cmark&A fly landed on the empty, staring vizor, and \boldcy{crawled} across it. & - \\
\bottomrule
\end{tabularx}
\end{center}\vspace{-8pt}
\caption{Some examples from the VUA verb classification development set. Metaphorical usages of the target word are bold faced, and literal usages are italicized. Leftmost columns show the correctness of prediction.}\vspace{-8pt}
\label{tab:examples}
\end{table*}

\paragraph{Error Analysis}
We sampled 100 errors of our best model from the VUA verb classification development set for analysis. Table~\ref{tab:examples} shows examples. 
Following the original annotation guideline,\footnote{http://www.vismet.org/metcor/documentation/home.html} we classify metaphors into five categories: direct metaphor, indirect metaphor, implicit metaphor, personification, and borderline case.  Indirect metaphor, the most common type for verbs, means that the basic meaning of a word is different from its contextual meaning.
Implicit metaphor occurs due to an underlying link which points to a recoverable metaphorical concept.

About half of the errors were false positives, and the other half were false negatives. Among the false negatives, 33\% are indirect metaphors, 18\% are personifications, and 2\% are direct metaphors. Among 55 false positives, 31\% of verbs have implicit arguments that are not explicitly mentioned in the context, 15\% have long range dependencies (at least five words away) from core arguments, 10\% have arguments with rare word senses, and 5\% have anthropomorphic arguments. Finally, we found about half of false negatives and 20\% of false positives to be borderline cases, showing the subjective nature of the task.

We sampled 257 dev examples that the CLS model gets wrong but the SEQ model gets correct. We found that the SEQ model outperforms the CLS model on detecting personifications, indirect metaphors, and direct metaphors involving uncommon verbs. 

%% file: related.tex
\section{Related Work}
There has been significant work on studying different features for metaphor detection, including concretenesss and abstractness~\cite{Turney2011LiteralAM,Tsvetkov2014MetaphorDW,koper2017improving}, imaginability~\cite{Broadwell2013UsingIA,Strzalkowski2013RobustEO}, feature norms~\cite{Bulat2017ModellingMW}, sensory features~\cite{Tekiroglu2015ExploringSF,Shutova:16BlackHA}, bag-of-words features~\cite{Kper2016DistinguishingLA}, and semantic class using WordNet~\cite{Hovy2013IdentifyingMW,Tsvetkov2014MetaphorDW}. More recently, embedding-based approaches~\cite{koper2017improving,Rei2017GraspingTF} showed gains on various benchmarks.

Many neural models with various features and architectures were introduced in the 2018 VUA Metaphor Detection Shared Task. They include LSTM-based models and CRFs augmented by linguistic features, such as WordNet, POS tags, concreteness score, unigrams, lemmas, verb clusters, and sentence-length manipulation~\cite{Swarnkar2018DiLSTMC, Pramanick2018AnLB, Mosolova2018ConditionalRF, Bizzoni2018BigramsAB, Wu2018NeuralMD}.
Researchers also studied different word embeddings, such as embeddings trained from corpora representing different levels of language mastery ~\cite{Stemle2018UsingLL} and binarized vectors that reflect the General Inquirer dictionary category of a word~\cite{Mykowiecka2018DetectingFW}. 
We show that contextualized word embedding significantly improves metaphor detection. We also study both sequence labeling and classification approaches, suggesting that sequence labeling approach enhances performance when  used to jointly predict the metaphoricity of all words in a sentence. 

%% file: conclusion.tex
\section{Conclusion}
In this paper, we present simple BiLSTM models augmented with contextualized word representation for metaphor detection. Our models establish new state-of-the-arts across multiple existing benchmarks, and our error analysis shows remaining challenges for metaphor detection. 